\documentclass[11pt,abstract=true]{scrartcl}

\usepackage[margin=1in]{geometry}
\usepackage{mathtools}
\usepackage{amsthm}
\usepackage{amsfonts}
\usepackage{amssymb}
\usepackage{booktabs}
\usepackage{xcolor}
\usepackage{xstring}
\definecolor{rwth-blue-100}{cmyk/RGB/HTML/gray}{1,.50,0,0/0,84,159/00549F/1}
\definecolor{rwth-blue-75}{cmyk/RGB/HTML/gray}{.75,.38,0,0/64,127,183/407FB7/.75}%
\definecolor{rwth-blue-50}{cmyk/RGB/HTML/gray}{.45,.14,0,0/142,186,229/8EBAE5/.5}
\definecolor{rwth-blue-25}{cmyk/RGB/HTML/gray}{.23,.07,0,0/199,221,242/C7DDF2/.25}%
\definecolor{rwth-blue-10}{cmyk/RGB/HTML/gray}{.09,.03,0,0/232,241,250/E8F1FA/.1}%
\definecolor{rwth-black-100}{cmyk/RGB/HTML}{0,0,0,1/0,0,0/000000}%
\definecolor{rwth-black-75}{cmyk/RGB/HTML}{0,0,0,.75/100,101,103/646567}%
\definecolor{rwth-black-50}{cmyk/RGB/HTML}{0,0,0,.5/156,158,159/9C9E9F}%
\definecolor{rwth-black-25}{cmyk/RGB/HTML}{0,0,0,.25/207,209,210/CFD1D2}%
\definecolor{rwth-black-10}{cmyk/RGB/HTML}{0,0,0,.1/236,237,237/ECEDED}%
\definecolor{rwth-magenta-100}{cmyk/RGB/HTML}{0,1,.25,0/227,0,102/E30066}%
\definecolor{rwth-magenta-75}{cmyk/RGB/HTML}{0,.75,.19,0/233,96,136/E96088}%
\definecolor{rwth-magenta-50}{cmyk/RGB/HTML}{0,.5,.13,0/241,158,177/F19EB1}%
\definecolor{rwth-magenta-25}{cmyk/RGB/HTML}{0,.25,.06,0/249,210,218/F9D2DA}%
\definecolor{rwth-magenta-10}{cmyk/RGB/HTML}{0,.10,.03,.02/253,238,240/FDEEF0}%
\definecolor{rwth-yellow-100}{cmyk/RGB/HTML}{0,0,1,0/255,237,0/FFED00}%
\definecolor{rwth-yellow-75}{cmyk/RGB/HTML}{0,0,.75,0/255,240,85/FFF055}%
\definecolor{rwth-yellow-50}{cmyk/RGB/HTML}{0,0,.50,0/255,245,155/FFF59B}%
\definecolor{rwth-yellow-25}{cmyk/RGB/HTML}{0,0,.25,0/255,250,209/FFFAD1}%
\definecolor{rwth-yellow-10}{cmyk/RGB/HTML}{0,0,.10,0/255,253,238/FFFDEE}%
\definecolor{rwth-petrol-100}{cmyk/RGB/HTML}{1,.3,.5,.3/0,97,101/006165}%
\definecolor{rwth-petrol-75}{cmyk/RGB/HTML}{.75,.23,.38,.23/45,127,131/2D7F83}%
\definecolor{rwth-petrol-50}{cmyk/RGB/HTML}{.5,.15,.25,.15/125,164,167/7DA4A7}%
\definecolor{rwth-petrol-25}{cmyk/RGB/HTML}{.25,.08,.13,.08/191,208,209/BFD0D1}%
\definecolor{rwth-petrol-10}{cmyk/RGB/HTML}{.1,.03,.05,.03/230,236,236/E6ECEC}%
\definecolor{rwth-turquoise-100}{cmyk/RGB/HTML}{1,0,.4,0/0,152,161/0098A1}%
\definecolor{rwth-turquoise-75}{cmyk/RGB/HTML}{.75,0,.3,.0/0,177,183/00B1B7}%
\definecolor{rwth-turquoise-50}{cmyk/RGB/HTML}{.5,0,.2,0/137,204,207/89CCCF}%
\definecolor{rwth-turquoise-25}{cmyk/RGB/HTML}{.25,0,.1,0/202,231,231/CAE7E7}%
\definecolor{rwth-turquoise-10}{cmyk/RGB/HTML}{.1,0,.04,0/235,246,246/EBF6F6}%
\definecolor{rwth-green-100}{cmyk/RGB/HTML}{.7,0,1,0/87,171,39/57AB27}%
\definecolor{rwth-green-75}{cmyk/RGB/HTML}{.52,0,.75,0/141,192,96/8DC060}%
\definecolor{rwth-green-50}{cmyk/RGB/HTML}{.35,0,.5,0/184,214,152/B8D698}%
\definecolor{rwth-green-25}{cmyk/RGB/HTML}{.18,0,.25,0/221,235,206/DDEBCE}%
\definecolor{rwth-green-10}{cmyk/RGB/HTML}{.07,0,.1,0/242,247,236/F2F7EC}%
\definecolor{rwth-maygreen-100}{cmyk/RGB/HTML}{.35,0,1,0/189,205,0/BDCD00}%
\definecolor{rwth-maygreen-75}{cmyk/RGB/HTML}{.26,0,.75,0/208,217,92/D0D95C}%
\definecolor{rwth-maygreen-50}{cmyk/RGB/HTML}{.18,0,.5,0/224,230,154/E0E69A}%
\definecolor{rwth-maygreen-25}{cmyk/RGB/HTML}{.09,0,.25,0/240,243,208/F0F3D0}%
\definecolor{rwth-maygreen-10}{cmyk/RGB/HTML}{.04,0,.1,0/249,250,237/F9FAED}%
\definecolor{rwth-orange-100}{cmyk/RGB/HTML}{0,.4,1,0/246,168,0/F6A800}%
\definecolor{rwth-orange-75}{cmyk/RGB/HTML}{0,.3,.75,0/250,190,80/FABE50}%
\definecolor{rwth-orange-50}{cmyk/RGB/HTML}{0,.2,.5,0/253,212,143/FDD48F}%
\definecolor{rwth-orange-25}{cmyk/RGB/HTML}{0,.1,.25,0/254,234,201/FEEAC9}%
\definecolor{rwth-orange-10}{cmyk/RGB/HTML}{0,.04,.1,0/255,247,234/FFF7EA}%
\definecolor{rwth-red-100}{cmyk/RGB/HTML}{.15,1,1,0/204,7,30/CC071E}%
\definecolor{rwth-red-75}{cmyk/RGB/HTML}{.11,.75,.75,0/216,92,65/D85C41}%
\definecolor{rwth-red-50}{cmyk/RGB/HTML}{0,.35,.47,.1/230,150,121/E69679}%
\definecolor{rwth-red-25}{cmyk/RGB/HTML}{0,.16,.23,.05/243,205,187/F3CDBB}%
\definecolor{rwth-red-10}{cmyk/RGB/HTML}{0,.06,.09,.02/250,235,227/FAEBE3}%
\definecolor{rwth-bordeaux-100}{cmyk/RGB/HTML}{.25,1,.70,.20/161,16,53/A11035}%
\definecolor{rwth-bordeaux-75}{cmyk/RGB/HTML}{.19,.75,.52,15/182,82,86/B65256}%
\definecolor{rwth-bordeaux-50}{cmyk/RGB/HTML}{.13,.5,.35,.1/205,139,135/CD8B87}%
\definecolor{rwth-bordeaux-25}{cmyk/RGB/HTML}{.06,.25,.18,.05/229,197,192/E5C5C0}%
\definecolor{rwth-bordeaux-10}{cmyk/RGB/HTML}{.03,.1,.07,.02/245,232,229/F5E8E5}%
\definecolor{rwth-violet-100}{cmyk/RGB/HTML}{.70,1,.35,.15/97,33,88/612158}%
\definecolor{rwth-violet-75}{cmyk/RGB/HTML}{.52,.75,.26,.11/131,78,117/834E75}%
\definecolor{rwth-violet-50}{cmyk/RGB/HTML}{.35,.5,.18,.08/168,133,158/A8859E}%
\definecolor{rwth-violet-25}{cmyk/RGB/HTML}{.18,.25,.09,.04/210,192,205/D2C0CD}%
\definecolor{rwth-violet-10}{cmyk/RGB/HTML}{.07,.1,.04,.02/237,229,234/EDE5EA}%
\definecolor{rwth-purple-100}{cmyk/RGB/HTML}{.6,.6,0,0/122,111,172/7A6FAC}%
\definecolor{rwth-purple-75}{cmyk/RGB/HTML}{.45,.45,0,0/155,145,193/9B91C1}%
\definecolor{rwth-purple-50}{cmyk/RGB/HTML}{.3,.3,0,0/188,181,215/BCB5D7}%
\definecolor{rwth-purple-25}{cmyk/RGB/HTML}{.15,.15,0,0/222,218,235/DEDAEB}%
\definecolor{rwth-purple-10}{cmyk/RGB/HTML}{.06,.06,0,0/242,240,247/F2F0F7}%

\usepackage{orcidlink}
\usepackage{doi}
\usepackage[numbers,sort&compress]{natbib}
\usepackage[automark]{scrlayer-scrpage}

\usepackage{algorithm}
\usepackage{algorithmic}
\usepackage{subcaption}
\usepackage{siunitx}
\usepackage{wrapfig}
\usepackage[inline]{enumitem}

\usepackage[capitalize]{cleveref}

\newtheorem{definition}{Definition}
\renewcommand{\vec}[1]{\boldsymbol{#1}}
\newcommand{\mat}[1]{\boldsymbol{#1}}
\DeclareMathOperator{\enc}{enc}
\DeclareMathOperator{\dec}{dec}

\usepackage{tikz}
\usepackage{tikz-3dplot}
\usepackage{pgfplots}
\usepackage{pgfplotstable}
\usetikzlibrary{arrows.meta, backgrounds, patterns, shapes.geometric}
\usepgfplotslibrary{groupplots}
\pgfplotsset{
    compat=1.18,
    discard if not/.style 2 args={
        filter discard warning=false,
        x filter/.code={
            \edef\tempa{\thisrow{#1}}
            \edef\tempb{#2}
            \ifx\tempa\tempb%
            \else
            
            \fi
        }
    }
}

\title{TopoEmbedX: A General Framework for Representation Learning on Topological Domains}

\author{%
    Florian Frantzen\textsuperscript{1}\,\orcidlink{0000-0003-0187-3738} \quad
    Ibrahem AlJabea\textsuperscript{2}\,\orcidlink{0000-0003-0038-844X} \quad
    Ines Henriques-Cadby\textsuperscript{3}\,\orcidlink{0000-0003-3916-7556} \\[0.3em]
    Theodore Papamarkou\textsuperscript{4,5}\,\orcidlink{0000-0002-9689-543X} \quad
    Mustafa Hajij\textsuperscript{6}\,\orcidlink{0000-0002-2625-9286} \quad
    Michael T. Schaub\textsuperscript{1}\,\orcidlink{0000-0003-2426-6404} \\[0.8em]
    \small
    \begin{tabular}{c@{\hspace{2em}}c}
        \textsuperscript{1}RWTH Aachen University       & \textsuperscript{2}Louisiana State University \\
        \textsuperscript{3}The University of Manchester & \textsuperscript{4}National Technical University of Athens \\
        \textsuperscript{5}PolyShape                    & \textsuperscript{6}University of San Francisco
    \end{tabular}
}

\date{}

\begin{document}

\maketitle
\begingroup
\deffootnote{0em}{0em}{}%
\renewcommand{\thefootnote}{}%
\footnotetext{%
    \begin{tabular}{@{}ll@{}}
        Coresponding: & \texttt{frantzen@netsci.rwth-aachen.de}, \texttt{ialjab2@lsu.edu}. \\
        Emails:       & \texttt{ines.henriques-cadby@manchester.ac.uk}, \texttt{tpapamarkou@mail.ntua.gr}, \texttt{mhajij@usfca.edu}, \\
                      & \texttt{schaub@netsci.rwth-aachen.de}
    \end{tabular}
}%
\endgroup

\begin{abstract}
    Topological structures such as simplicial complexes, hypergraphs, and cell complexes extend standard graph models by modeling higher-order relationships.
    These structures appear in many modern datasets and require specialized methods for generating meaningful embeddings.
    In this paper, we introduce \texttt{TopoEmbedX}, a unified framework for embedding a wide range of topological domains into Euclidean spaces.
    The package brings together several existing topological embedding algorithms---\texttt{DeepCell}, \texttt{Cell2Vec}, \texttt{CellDiff2Vec}, \texttt{HOLE}, and \texttt{HOGLEE}---and introduces five new algorithms: \texttt{ComplexNetMF}, \texttt{ComplexRep}, \texttt{ComplexRandNE}, \texttt{ComplexWalklets}, and \texttt{ComplexHeat}.
    These algorithms extend well-known graph embedding techniques to higher-order settings using the augmented Hasse graph of a topological domain.
    \texttt{TopoEmbedX} provides a clear, consistent, and easy-to-use framework for topological representation learning.
    Experiments show that the embeddings generated by \texttt{TopoEmbedX} support tasks such as classification and regression across multidimensional data.
\end{abstract}

\section{Introduction}%
\label{section:introduction}

Graph-based network models have become a cornerstone of modern data analysis, effectively capturing relationships between entities across a wide range of domains, including biological and chemical structures \citep{Li:2022, Reiser:2022,Wee:2025}, social systems \citep{Hamilton:2022}, communication networks~\citep{SuarezVarela:2023, Abadal:2022}, and transportation infrastructure \citep{Barabasi:2013}.
These models support complex analytical tasks such as epidemic modeling \citep{Harko:2014}, link-based ranking \citep{Page:1999}, and path finding \citep{Madkour:2017}.
To accommodate richer structures, researchers have extended basic network models with node and edge attributes (weights, labels, timestamps), to study heterogeneous networks \citep{Yang:2022}, and dynamic properties \citep{Rossetti:2019}.
Real-world systems also frequently exhibit multi-entity interactions that are not always well represented by the binary paradigm offered by graphs \citep{Granovetter:1973, Bavelas:1950, Ugander:2013}.
Higher-order networks address this limitation by generalizing beyond pairwise connectivity.

Representation learning offers a principled approach to encoding network structure into fixed-length vector embeddings, enabling downstream tasks such as node classification and link prediction.
Early work by \citet{Rossi:2018} demonstrated the value of incorporating motifs into learned embeddings, and subsequent methods further leveraged supernodes and random walks to preserve higher-order structural similarity \citep{Shao:2022}. Despite these advances, existing approaches remain largely confined to graph-based representations. Systematic methods for embedding higher-order networks across all topological dimensions remain underdeveloped \citep{Vaida:2019, Gong:2023}.
Topological Representation Learning (TRL) addresses this challenge by learning embeddings of topological domains in Euclidean space while preserving their intrinsic structural relationships. TRL has demonstrated strong results in graph representation learning \citep{Cui:2019, Narayanan:2017, Tsitsulin:2018}, with successful applications spanning node classification, whole-graph classification, link prediction, and similarity analysis.
Recent work has extended TRL to richer topological domains---including simplicial complexes, hypergraphs, and cell complexes---enabling the joint embedding of cells across multiple dimensions \citep{Hajij:2020}.
TRL methods have further been applied to image segmentation \citep{Hu:2021, Gupta:2023}, shape generation \citep{Waibel:2022}, cell embeddings via random walks \citep{Billings:2019, Schaub:2020}, and persistent homology regularization \citep{Chen:2019}.
Nevertheless, a general-purpose tool capable of systematically embedding arbitrary topological domains across all dimensional levels remains absent from the literature.

In this paper, we address this gap by extending the \texttt{TopoX} framework~\citep{Hajij:2024}, which includes a preliminary variant of \texttt{TopoEmbedX} consisting primarily of five existing topological embedding algorithms.
The present work substantially extends that preliminary version and establishes \texttt{TopoEmbedX} as a standalone framework for topological embedding.
To this end, we introduce and implement five new topological embedding algorithms and develop a unified, open-source framework that generalizes classical graph embedding algorithms to higher-order topological domains—including simplicial complexes, hypergraphs, cell complexes, and combinatorial complexes (CCs).
\texttt{TopoEmbedX} thus enables systematic, multidimensional topological representation learning with consistent notation, user-friendly descriptions, and comprehensive experimental validation, extending the graph embedding paradigm to a broad class of topological domains.
In this paper, an embedding algorithm takes a (finite) topological domain together with modeling choices such as neighborhood operators, embedding dimension, and method-specific parameters, and returns vector representations for the selected cells.
\texttt{TopoEmbedX} realizes this workflow by constructing an augmented Hasse graph whose nodes are cells and whose edges encode the neighborhood relations used by the chosen graph embedding backend.

\paragraph{Contributions.}
We present the following contributions in this paper:
\begin{enumerate}
    \item
        \emph{Formalization and generalization of topological representation learning.}
        We formalize the problem of representation learning on topological domains and systematically extend classical graph embedding techniques---including random walks, diffusion processes, heat kernels, randomized projections, and matrix factorization---to higher-order topological structures.
        This generalization is grounded in the Hasse graph, which serves as a principled and unified structural description of higher-order models across all topological dimensions.
    \item
        \emph{A unified algorithmic framework integrating established and novel techniques.}
        We implement five existing embedding algorithms (\texttt{DeepCell}, \texttt{Cell2Vec}, \texttt{CellDiff2Vec}, \texttt{HOLE}, \texttt{HOGLEE}) and introduce five novel algorithms (\texttt{ComplexNetMF}, \texttt{ComplexRep}, \texttt{ComplexRandNE}, \texttt{ComplexWalklets}, \texttt{ComplexHeat}), each generalizing a well-established graph embedding technique to topological domains via Hasse-based incidence operators.
        These ten algorithms form a coherent, extensible, and theoretically grounded algorithmic foundation for topological representation learning.

    \item
        \emph{Open-source software with comprehensive experimental validation.}
        We release \texttt{TopoEmbedX} as an open-source Python package featuring efficient data structures and a clean, well-documented API designed for accessibility and extensibility.
        The framework is rigorously evaluated through experiments on multidimensional topological datasets, benchmarking performance across two supervised edge-level tasks: edge classification and edge regression.
\end{enumerate}

\paragraph{Paper structure.}
The remainder of this paper is organized as follows.
\Cref{section:topological-domain} introduces topological domains and the neighborhood operators used throughout the paper.
\Cref{section:augmented-hasse-graph} develops the notion of a Hasse graph and the augmented Hasse graph, which provide the structural backbone for our generalization strategy.
\Cref{section:representation-learning} formulates higher-order representation learning and explains how it reduces to graph representation learning on augmented Hasse graphs.
\Cref{section:topoembedx-overview} then presents \texttt{TopoEmbedX} and its algorithmic components.
\Cref{section:experiments} summarizes the experimental protocol and evaluation tasks.
Finally, \cref{section:conclusion} concludes by outlining directions for future work.

\section{Overview of Topological Domains}%
\label{section:topological-domain}
Representation learning on higher-order data requires domains that can encode more than pairwise relations.
Hypergraphs naturally model set-valued interactions, while simplicial and cell complexes encode hierarchical incidence relations between cells of different dimensions. CCs~\cite{Hajij:2023} provide a common abstraction that can represent both types of structures: arbitrary set-valued cells together with a rank function that is compatible with inclusion.

In this paper, we use the term \emph{topological domain} to refer to the input object on which the embedding algorithms in \texttt{TopoEmbedX} are defined. This object is modeled as a CC \cite{Hajij:2023}: a ranked collection of cells equipped with an inclusion relation. This terminology is intentional. The phrase \emph{topological domain} emphasizes the algorithmic role of the input as a general domain for representation learning, while the CC provides the precise mathematical model.
This choice allows \texttt{TopoEmbedX} to treat embedding algorithms independently of the specific input type, whether the data are given as a graph, simplicial complex, hypergraph, cell complex, or CC.

\begin{definition}[Topological domain]
    \label{def:topological_domain}
    A \emph{topological domain} $\mathcal{D}$ is a CC defined by a triple $(\mathcal{V}, \mathcal{X}, \operatorname{rk})$, where
    \begin{enumerate}
        \item $\mathcal{V}$ is a finite nonempty set of vertices;
        \item $\mathcal{X} \subseteq \mathcal{P}(\mathcal{V}) \setminus \{\emptyset\}$ is a finite collection of nonempty subsets of $\mathcal{V}$ such that $\{v\} \in \mathcal{X}$ for every $v \in \mathcal{V}$;
        \item $\operatorname{rk} \colon \mathcal{X} \to \mathbb{Z}_{\geq 0}$ is a rank function satisfying $\operatorname{rk}(\{v\}) = 0$ for every $v \in \mathcal{V}$, and given $x,y \in \mathcal{X}$, $\operatorname{rk}(x) \leq \operatorname{rk}(y)$, whenever $x\subseteq y$.
    \end{enumerate}
\end{definition}

For ease of notation, we write $\mathcal{D}$ simply as $\mathcal{X}$. Each element $x\in \mathcal{X}$ is said to have rank $\operatorname{rk}(x)$, and $\mathcal{X}$ is said to have dimension $\dim(\mathcal{X}) = \max_{x \in \mathcal{X}} \operatorname{rk}(x)$. We refer to elements of $\mathcal{X}$ as \emph{cells}, and those of rank $k$, as  $k$-\emph{cells}, with  $\mathcal{X}^{k}$ denoting the set of all $k$-cells (i.e. $\mathcal{X}^k=\operatorname{rk}^{-1}(k)\,$).
A cell $x \in \mathcal{X}$ is said to be a \emph{face} of $y \in \mathcal{X}$, and  $y$ a \emph{coface} of $x$, whenever $x \subsetneq y$.
Finally, we recall the \emph{cover relation}. A cell $x$ is said to be a \textit{lower cover of } a cell $y$ , and $y$ an \textit{upper cover} of $x$, denoted $x\prec y$, if $x \subsetneq y$ and there exists no cell $z \in \mathcal{X}$ such that $x \subsetneq z \subsetneq y$.

The definition of topological domain in \cref{def:topological_domain}  provides a unified framework encompassing graphs, simplicial complexes, hypergraphs, and cell complexes as special cases. In a graph, vertices and edges are assigned ranks $0$ and $1$ respectively. In a simplicial complex, every simplex $\sigma$ satisfies $\operatorname{rk}(\sigma) = |\sigma| - 1$, and the downward-closure property ensures that all faces of every simplex are present. In a hypergraph, hyperedges are arbitrary nonempty subsets of vertices, with rank assignments that respect the rank--the inclusion relation of \cref{def:topological_domain}, but without requiring downward closure. In a cell complex, each cell carries an explicit dimension that coincides with its rank, and the boundary structure of cells is encoded by the rank--inclusion relation, recovering the standard notion of a CW complex.
The framework therefore captures both hierarchical inclusion relations and general set-based interactions \cite{Hajij:2023}.

\subsection{Neighborhood Structures}%
\label{section:neighborhood-function}

A fundamental component in topological representation learning is the notion of neighborhoods, which formalize local interactions within the topological domain $\mathcal{X}$.
\begin{definition}[Neighborhood Function]
    A neighborhood function on $\mathcal{X}$ is a map $\mathcal{N} \colon \mathcal{X} \to \mathcal{P}(\mathcal{X})$, which assigns to each cell $x$ in $\mathcal{X}$ a collection of neighbor cells $\mathcal{N}(x) \subseteq \mathcal{X}$, referred to as \emph{the neighborhood} of $x$.
\end{definition}
This formalism is essential for defining and analyzing local dependencies, particularly in machine learning models that exploit hierarchical and relational structures. To illustrate it, we introduce \emph{incidence neighborhoods} and two commonly used same-rank neighborhood functions.

Given a cell $x \in \mathcal{X}$, its \emph{incidence neighborhood}  is defined as
\begin{align}
    \mathcal{N}_{\mathrm{inc}}(x) = \left\{ y \in \mathcal{X} \mid y \prec x \right\},
\end{align}
which consists of the lower covers of $x$.

The \emph{adjacency neighborhood} of a cell $x \in \mathcal{X}^k$ is defined by
\begin{align}
    \mathcal{N}_{\mathrm{adj}}(x) = \left\{ y \in \mathcal{X}^k \setminus \{x\} \mid x \prec z \text{ and } y \prec z , \text{ for some } z \in \mathcal{X} \right\},
\end{align}
thereby capturing cells that share a common coface.

The \emph{co-adjacency neighborhood} of a cell $x \in \mathcal{X}^k$ is defined by
\begin{align}
    \mathcal{N}_{\mathrm{coadj}}(x) = \left\{ y \in \mathcal{X}^k \setminus \{x\} \mid z \prec x \text{ and } z \prec y, \text{ for some }z \in \mathcal{X} \right\},
\end{align}
thereby capturing cells that share a common face.

For example, two edges contained in a common $2$-cell are adjacent, whereas two edges sharing a vertex are co-adjacent. Ordinary graph adjacency between vertices is therefore recovered as a rank-$0$ adjacency relation in this convention ($k=0$).

In practice, neighborhood functions are stored via sparse matrices called \emph{neighborhood matrices}.

\begin{definition}[Neighborhood Matrix]
    Let $\mathcal{N}$ be a neighborhood function on a topological domain $\mathcal{X}$, and  $\mathcal{Y} = \{ y_1, \ldots, y_n \} , \ \mathcal{Z} = \{ z_1, \ldots, z_m \} \subseteq \mathcal{X}$ be two collections of cells satisfying $\mathcal{N}(\mathcal{Y}) \subseteq \mathcal{Z} $.
    The \emph{neighborhood matrix} associated with $\mathcal{N}\big|_\mathcal{Y}$ is defined as the $m\times n$ matrix whose entries are
    \begin{align}
        [\mat{N}]_{ij} & =
        \begin{cases}
            1 & \text{if } z_i \in \mathcal{N}(y_j), \\
            0 & \text{otherwise}.
        \end{cases}
        \text{ for } i=1,\dots,m, j=1,\dots,n
    \end{align}
\end{definition}
For incidence neighborhoods the matrix is rectangular, reflecting the cross-rank nature of the relation.
For adjacency neighborhoods, the matrix is square (same-rank), and is symmetric if the adjacency relation is symmetric.
The classical graph adjacency matrix and incidence matrix are both special cases of neighborhood matrices when $\mathcal{X}$ is a graph.
These matrices encode the local neighborhood relationships between cells, generalizing both the adjacency matrix and the incidence matrix of a graph to topological domains of arbitrary dimension.
In practice, the rich structure of a topological domain, with its multiple ranks and both incidence and adjacency relations, means that a collection of neighborhood functions $\mathsf{N}=\{\mathcal{N}_1,\dots,\mathcal{N}_k\}$ is needed simultaneously to fully describe the local environment of a cell.

\section{Augmented Hasse Graph of a Topological Domain}%
\label{section:augmented-hasse-graph}

In this section, we introduce the notion of an \textit{augmented Hasse graph}, an extension of the classical Hasse graph that incorporates additional neighborhood relationships between cells beyond standard incidence, enabling topological domains to be processed by graph-based embedding methods as defined in section \cref{section:representation-learning}.
We revisit the concept of a Hasse graph and extend it to that of an \textit{augmented Hasse graph}. The latter represents a topological domain as a graph, enabling the use of graph-based techniques for representation learning.
A key conceptual distinction exists between the two graph structures associated with a topological domain: the \emph{canonical} Hasse graph depends only on the incidence relations of the domain and requires no additional choices, whereas the \textit{augmented} Hasse graph additionally incorporates neighborhood relationships chosen for a the specific representation learning task, and is therefore \emph{model-dependent}.

\paragraph{Role in \texttt{TopoEmbedX}}

In \texttt{TopoEmbedX}, the augmented Hasse graph serves as the core representation. A topological domain $\mathcal{X}$ equipped with a neighborhood collection $\mathsf{N}$ is mapped to the derived augmented Hasse graph $\mathcal{H}_{\mathcal{X}}(\mathsf{N})$, after which a standard graph embedding algorithm is applied.
The node--cell correspondence between $\mathcal{H}_{\mathcal{X}}(\mathsf{N})$ and $\mathcal{X}$ then yields embeddings for the original topological domain without loss of cell identity.
This reduction is principled, as it preserves all relations encoded by $\mathsf{N}$, and modular, as changing the graph embedding algorithm induces a different topological embedding method while leaving the reduction stage unchanged.
This design yields the unified and extensible framework described in \cref{section:topoembedx-overview}.

\subsection{Hasse Graph of a Topological Domain}
\label{section:hasse-graph}

Let $\mathcal{X}$ be a topological domain.
Because each cell $x \in \mathcal{X}$ is a nonempty subset of the vertex set $\mathcal{V}$, set inclusion induces a partial order on $\mathcal{X}$. Hence, $(\mathcal{X}, \subseteq)$ is a finite poset, which we call the \emph{containment poset}.
The Hasse graph encodes the inclusion relations of this containment poset as follows.
\begin{definition}[Hasse Graph]
    Let $\mathcal{X}$ be a topological domain equipped with the inclusion poset.
    The \emph{Hasse graph} $\mathcal{H}_\mathcal{X}$ is the directed graph with node set
    $\mathcal{V}(\mathcal{H}_\mathcal{X}) = \mathcal{X}$ and a directed edge from $x$ to $y$
    if and only if $x \prec y$.
\end{definition}
The Hasse graph is \emph{canonical} in that it depends only on the containment structure of $\mathcal{X}$ and requires no additional modeling choices.
Its directed edges encode cover relations between cells.
In particular, the full containment order on $\mathcal{X}$ is recoverable from the Hasse graph: $x \subseteq y$ if and only if there exists a directed path from $x$ to $y$ in $\mathcal{H}_\mathcal{X}$, so no inclusion information is lost in passing from the poset to its Hasse graph.

An important special case occurs when the poset admits a \emph{graded structure}, i.e., when there exists a rank function $\operatorname{rk} \colon \mathcal{X} \to \mathbb{Z}_{\geq 0}$ such that $\operatorname{rk}(y) = \operatorname{rk}(x) + 1$ for every cover relation $x \prec y$. This property holds for simplicial complexes, where $\operatorname{rk}(\sigma) = |\sigma| - 1$, and for regular cell complexes, where rank coincides with cell dimension.
Since the Hasse graph contains exactly the cover relations of the containment poset, the graded structure induces a layered decomposition by rank, with every edge connecting a cell of rank $k$ to one of rank $k+1$. This makes the Hasse graph a natural computational representation for simplicial and regular cell complexes.

\paragraph{\textbf{Why Go Beyond the Hasse Graph?}}
The Hasse graph provides a complete representation of the containment structure of a topological domain, since the full containment order can be recovered from directed paths in  $\mathcal{H}_\mathcal{X}$.
Although augmentation is not necessary to represent the domain itself, embedding algorithms typically operate on an explicit computational graph, whose edges determine random-walk transitions, message-passing neighborhoods, diffusion pathways, and spectral operators. Many relations relevant to an embedding task are computable from the Hasse graph, yet they do not appear as explicit edges. For example, two edges in a simplicial complex that share a vertex are co-adjacent; this relation arises naturally in the modeling of signal flow and physical connectivity~\citep{Schaub:2021,Schaub:2022,Barbarossa:2020, Barbarossa:2020b, Hernandezserrano:2020}, yet this relation is encoded only indirectly, through their shared lower-rank face. Making such relations explicit modifies the inductive bias of the resulting embedding method: it can shorten relevant paths, alter transition probabilities, and determine which cells exchange information in a message-passing layer. We refer to these algorithm- or task-specific choices as \emph{task-specific neighborhood structure}. This motivates the augmented Hasse graph, which preserves the canonical Hasse graph while admitting additional, task-specific neighborhood relations as explicit edges.

\subsection{Augmented Hasse Graph}%
\label{section:augmented-hasse-graph-def}

The augmented Hasse graph extends the classical Hasse graph by incorporating additional inter-cell relationships induced by a chosen collection of neighborhood functions.
These added edges encode interactions that may be relevant to a given learning task but are not represented as single edges in the canonical Hasse graph.
Examples include adjacency through shared cofaces, co-adjacency through shared lower covers, and higher-order proximity relations.

\begin{figure}[t]
    \centering
    \resizebox{\textwidth}{!}{%
        \input{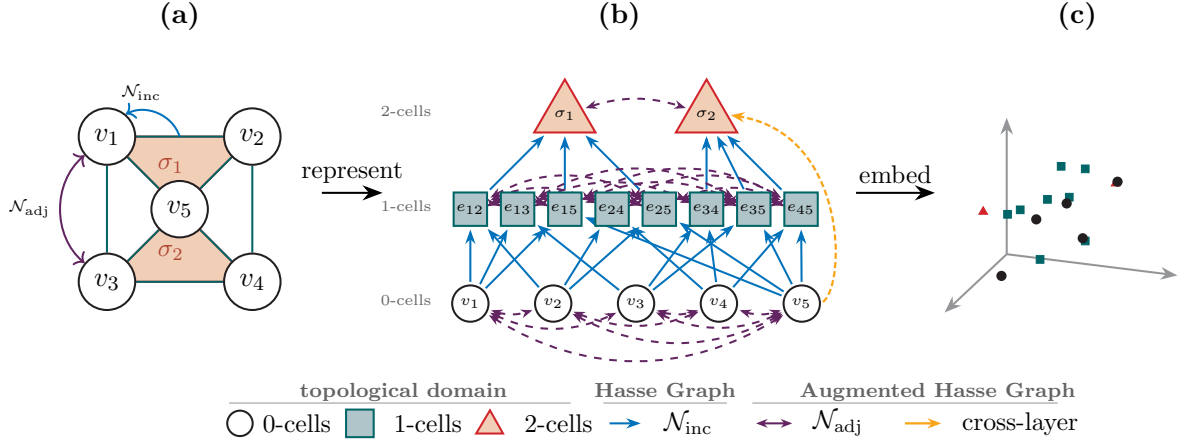}
    }
    \caption{%
        Example of a topological representation learning algorithm that embeds a topological domain into a three-dimensional vector space.
        (a) Topological domain $\mathcal{X}$. (b) Augmented Hasse graph $\mathcal{H}_{\mathcal{X}}(\mathsf{N})$ with cover edges, same-rank edges induced by $\mathcal{N}_{\mathrm{adj}}$, and an example of a cross-layer edge between $0$-cells and $2$-cells. (c) Embedding space $\mathbb{R}^3$.
    }%
    \label{fig:example}
\end{figure}

Let $\mathcal{X}$ be a topological domain with Hasse graph $\mathcal{H}_\mathcal{X}$, and let $\mathsf{N} = \{\mathcal{N}_1, \ldots, \mathcal{N}_n\}$ be a collection of neighborhood functions on $\mathcal{X}$ in the sense of \cref{section:neighborhood-function}.
We say that an \emph{augmented edge} from $y$ to $x$ is induced by $\mathsf{N}$ if there exists some $\mathcal{N}_i \in \mathsf{N}$ such that $y \in \mathcal{N}_i(x)$.
With this convention, $\mathcal{N}_{\mathrm{inc}}(x)$ lists lower covers of $x$, while the corresponding directed edges point upward from those lower covers to $x$, matching the orientation of the canonical Hasse graph.

\begin{definition}[Augmented Hasse Graph]
    Let $\mathcal{X}$ be a topological domain, and let $\mathsf{N} = \{\mathcal{N}_1, \ldots, \mathcal{N}_n\}$ be a collection of neighborhood functions on $\mathcal{X}$.
    The \emph{augmented Hasse graph} of $\mathcal{X}$ with respect to $\mathsf{N}$ is the directed graph with node set $\mathcal{V}(\mathcal{H}_{\mathcal{X}}(\mathsf{N})) = \mathcal{X}$ and edge set
    \begin{align}
        \mathcal{E}(\mathcal{H}_{\mathcal{X}}(\mathsf{N}))
        & = \mathcal{E}(\mathcal{H}_{\mathcal{X}}) \cup \mathcal{E}_{\mathsf{N}},
    \end{align}
    where
    \begin{align}
        \mathcal{E}_{\mathsf{N}}
        & = \{
            (y, x) \in \mathcal{X} \times \mathcal{X} \mid x \neq y,\,
        \exists \mathcal{N}_i \in \mathsf{N} \text{ such that } y \in \mathcal{N}_i(x)\}.
    \end{align}
\end{definition}
Note that the non-augmented edges in $\mathcal{E}(\mathcal{H}_{\mathcal{X}}(\mathsf{N}))$ are precisely the directed cover-relation edges inherited from the Hasse graph.
To keep the terminology unambiguous, we reserve the term \emph{augmented edge} for an edge induced by $\mathsf{N}$ that is not already present in $\mathcal{E}(\mathcal{H}_{\mathcal{X}})$.
Several remarks are in order regarding both the conceptual status and the practical utility of this construction.

\paragraph{Non-canonicity}
Unlike the classical Hasse graph, the augmented Hasse graph is \emph{not canonical}.
It depends on the neighborhood collection $\mathsf{N}$, whose specification is a modeling choice determined by the representation-learning task at hand.
Accordingly, different choices of $\mathsf{N}$ induce different augmented Hasse graphs on the same underlying topological domain.
This dependence allows $\mathcal{H}_{\mathcal{X}}(\mathsf{N})$ to be tailored to the relations most relevant to a given task, including adjacency, co-adjacency, higher-order incidence, or combinations thereof.
In \texttt{TopoEmbedX}, the choice of $\mathsf{N}$ is made by the user in the API of the embedding algorithm, each of which selects the neighborhood operators appropriate to its underlying graph embedding technique.

\paragraph{Construction}
From a computational perspective, the augmented Hasse graph is obtained by starting with the canonical Hasse graph and adding the non-canonical membership relations induced by the neighborhood functions in $\mathsf{N}$.
In implementations, each function $\mathcal{N}_i \in \mathsf{N}$ can be represented by its associated sparse neighborhood matrix $\mat{N}_i$.
Thus, $\mathcal{H}_{\mathcal{X}}(\mathsf{N})$ can be assembled efficiently by taking the union of the canonical Hasse edges with the sparse supports of these neighborhood matrices.

\paragraph{Recovering the classical Hasse graph}
The classical Hasse graph is recovered as the special case in which $\mathsf{N}$ is empty and no augmented edges are added.
More generally, $\mathcal{H}_\mathcal{X}$ is always a subgraph of $\mathcal{H}_{\mathcal{X}}(\mathsf{N})$: the augmentation can add edges but cannot remove existing ones.
Hence, the cover relations of the containment poset remain present in the augmented Hasse graph for any admissible choice of $\mathsf{N}$.

\paragraph{Scope of augmentation}
The neighborhood collection $\mathsf{N}$ may encode a broad class of relations.
For example, adjacency functions can relate distinct rank-$k$ cells that are faces of a common coface, thereby inducing horizontal edges within rank layers.
Co-adjacency functions induce edges between cells sharing a lower-dimensional face, and multi-hop incidence functions extend connectivity beyond immediate rank transitions.
In principle, any neighborhood function in the sense of \cref{section:neighborhood-function} can induce edges in the augmented Hasse graph, which renders the construction general.

\paragraph{Illustrative example}
Consider a simplicial complex $\mathcal{X}$ representing a two-dimensional surface, comprising $0$-cells (vertices), $1$-cells (edges), and $2$-cells (triangles).
The Hasse graph $\mathcal{H}_\mathcal{X}$ contains only directed edges from vertices to incident edges and from edges to incident triangles.
Suppose, however, that we augment this structure with the co-adjacency function on $1$-cells, which relates pairs of edges sharing a common vertex, and with the adjacency function on $0$-cells, which relates pairs of vertices connected by a common edge.
The resulting augmented Hasse graph $\mathcal{H}_{\mathcal{X}}(\mathsf{N})$ then contains additional horizontal edges between co-adjacent edges and between adjacent vertices.
These same-rank relations are not directly represented in the canonical Hasse graph, yet they may be essential for embedding techniques that rely on local structural similarity within a rank.
If one further includes proximity-based neighborhoods, for example by declaring two vertices to be neighbors whenever their Euclidean distance is below a threshold, the augmented Hasse graph also incorporates geometric information, thereby yielding a richer representation of the domain.

\section{Higher-order Representation Learning}%
\label{section:representation-learning}

Graph representation learning aims to map vertices, edges, or subgraphs to vectors in Euclidean space while preserving key structural properties of the graph.
Higher-order representation learning \cite{Hajij:2020,Bick:2023a} extends this principle to the topological setting by seeking embeddings of the cells of a topological domain that retain the domain's fundamental structural relationships in the resulting representation.

The augmented Hasse graph serves as the principal mechanism linking topological domains to classical graph-based embedding techniques.
Once a topological domain is represented as a graph whose vertices are cells and whose edges encode incidence together with selected neighborhood relations, standard graph embedding techniques can be applied in the higher-order setting.
Moreover, the correspondence between augmented Hasse graphs and CCs, established in \citet[Theorem 8.4]{Hajij:2023}, shows that graph-based deep learning constructions admit a natural generalization to the combinatorial setting.
Accordingly, we treat the augmented Hasse graph as the primary reduction device and interpret higher-order representation learning as graph representation learning performed on a structured graph derived from the original topological domain.

\paragraph{Single-rank formulation.}
Given a complex $\mathcal{X}$ and a fixed rank $k$, the representation learning task amounts to learning an encoder--decoder pair $(\enc, \dec)$, where the encoder map $\enc \colon \mathcal{X}^k \to \mathbb{R}^d$ assigns to each $k$-cell $x^k \in \mathcal{X}^k$ a vector representation that encodes its structural role relative to the remaining cells of $\mathcal{X}$, and the decoder map $\dec \colon \mathbb{R}^d \times \mathbb{R}^d \to \mathbb{R}$ assigns to each pair of cell embeddings a scalar measuring their degree of relatedness.
These components are optimized with respect to a task-specific similarity measure $\operatorname{sim} \colon \mathcal{X}^k \times \mathcal{X}^k \to \mathbb{R}$ by minimizing
\begin{equation}
    \mathcal{L}_k
    =
    \sum_{x^k \in \mathcal{X}^k}\sum_{y^k \in \mathcal{X}^k}
    \ell\!\left(
        \dec(\enc(x^k),\enc(y^k)),\,
        \operatorname{sim}(x^k,y^k)
    \right),
    \label{eq:single-rank-loss}
\end{equation}
where $\ell \colon \mathbb{R} \times \mathbb{R} \to \mathbb{R}$ is a prescribed loss function.
This formulation makes explicit that the learned representation should preserve the similarity notion induced by the chosen neighborhood, diffusion, spectral, or structural operator.
As established in \citet{Hajij:2023}, higher-order representation learning can be reduced to graph representation learning.

\paragraph{Multi-rank formulation.}
The single-rank objective in \cref{eq:single-rank-loss} captures structural relationships among cells of a fixed dimension $k$, but it does not explicitly model the cross-rank dependencies that are characteristic of a topological domain.
In many settings, the structural role of a $k$-cell is determined not only by its relations to other $k$-cells, but also by the $(k-1)$-cells on its boundary, the $(k+1)$-cells to which it belongs, and, more generally, its position within the entire hierarchical structure of $\mathcal{X}$.
Accounting for this multi-level context therefore requires extending the representation learning objective to operate on \emph{tuples of cells drawn from multiple ranks simultaneously}.

Let $\mathbf{k} = (k_1, \ldots, k_m)$ be a tuple of ranks, not necessarily distinct, with $0 \leq k_i \leq \dim(\mathcal{X})$ for each $i$.
For each rank $k_i$, let $\enc_i \colon \mathcal{X}^{k_i} \to \mathbb{R}^{d_i}$ be an encoder that maps $k_i$-cells to a $d_i$-dimensional Euclidean space.
The multi-rank decoder
\begin{equation}
    \dec \colon \mathbb{R}^{d_1} \times \cdots \times \mathbb{R}^{d_m} \to \mathbb{R}
\end{equation}
takes a tuple of cell embeddings drawn from potentially different ranks and returns a scalar quantifying their joint structural relatedness.
The corresponding multi-rank similarity measure
\begin{equation}
    \operatorname{sim} :
    \mathcal{X}^{k_1} \times \cdots \times \mathcal{X}^{k_m} \to \mathbb{R}
\end{equation}
assigns to each $m$-tuple of cells $(x^{k_1}, \ldots, x^{k_m})$ a scalar reflecting the extent to which those cells are structurally related across all $m$ rank levels.
The multi-rank representation learning objective is then
\begin{equation}
    \mathcal{L}_{\mathbf{k}}
    =
    \sum_{x^{k_1} \in \mathcal{X}^{k_1}}
    \cdots
    \sum_{x^{k_m} \in \mathcal{X}^{k_m}}
    \ell\!\left(
        \dec\!\left(
            \enc_1(x^{k_1}),\, \ldots,\, \enc_m(x^{k_m})
        \right),\,
        \operatorname{sim}(x^{k_1}, \ldots, x^{k_m})
    \right).
    \label{eq:multi-rank-loss}
\end{equation}
This objective generalizes \cref{eq:single-rank-loss} directly: setting $m = 2$ and $k_1 = k_2 = k$ recovers the single-rank pairwise objective.
The extension to $m > 2$ or to tuples with $k_i \neq k_j$ is both natural and computationally coherent, since the algebraic structure of the objective---summing over tuples, evaluating the decoder, comparing against similarity, and accumulating the loss---remains unchanged.

\paragraph{Multi-rank similarity via the augmented Hasse graph.}
The central modeling question is how to define $\operatorname{sim}(x^{k_1}, \ldots, x^{k_m})$.
The augmented Hasse graph $\mathcal{H}_{\mathcal{X}}(\mathsf{N})$ offers a natural answer.
Given a neighborhood collection $\mathsf{N}$, the augmented Hasse graph encodes all pairwise relationships between cells of the ranks deemed relevant for the task.
For the symmetric proximity convention used by the multi-rank embedding algorithms below, let $\mat{A}^{\mathrm{sym}}$ be the adjacency matrix of $\mathcal{H}^{\mathrm{sym}}_{\mathcal{X}}(\mathsf{N})$, let $\mat{D}$ be its degree matrix, and set $\mat{P}=\mat{D}^{-1}\mat{A}^{\mathrm{sym}}$.
Thus, $\mat{P}$ is the row-normalized transition matrix of a random walk on the symmetrized augmented Hasse graph.
The $t$-step transition matrix $\mat{P}^t$ thereby induces a family of pairwise similarity measures, where $[\mat{P}^t]_{x,y}$ denotes the probability of reaching cell $y$ from cell $x$ in $t$ steps while traversing cells of all intermediate ranks along the way.
For the multi-rank objective in \cref{eq:multi-rank-loss}, one natural choice is
\begin{equation}
    \operatorname{sim}(x^{k_1}, \ldots, x^{k_m})
    \;=\;
    \prod_{i=1}^{m-1} [\mat{P}^{t_i}]_{x^{k_i},\, x^{k_{i+1}}},
    \label{eq:chain-similarity}
\end{equation}
which measures the probability of traversing the chain $x^{k_1} \to x^{k_2} \to \cdots \to x^{k_m}$ through $\mathcal{H}^{\mathrm{sym}}_{\mathcal{X}}(\mathsf{N})$ via a sequence of walks of lengths $t_1, \ldots, t_{m-1}$, respectively.
This chain similarity is large when the cells are mutually reachable by short paths in $\mathcal{H}^{\mathrm{sym}}_{\mathcal{X}}(\mathsf{N})$, and it decreases as the cells become structurally more distant within the hierarchical organization of $\mathcal{X}$.
Other aggregations are also possible: one may replace the product with a minimum, a sum, or a learned aggregation, depending on whether the tuple similarity should emphasize the weakest link, cumulative proximity, or a task-adaptive combination of the pairwise similarities.

\paragraph{Relationship to the single-rank case and to the augmented Hasse
graph reduction.}
The multi-rank objective in \cref{eq:multi-rank-loss} is computationally analogous to the single-rank case, and it admits the same reduction to graph representation learning via the augmented Hasse graph.
In particular, because all cells of $\mathcal{X}$---independently of rank---appear as nodes in $\mathcal{H}_{\mathcal{X}}(\mathsf{N})$, any graph embedding method applied to $\mathcal{H}_{\mathcal{X}}(\mathsf{N})$ yields embeddings for cells of all ranks simultaneously.
The multi-rank loss in \cref{eq:multi-rank-loss} can therefore be evaluated directly using these joint embeddings, without any alteration to the underlying graph embedding algorithm.
A practical advantage of the augmented Hasse graph reduction is that it embeds all ranks in one Euclidean space using a single graph, so within-rank and cross-rank relations are learned jointly.
The runtime still depends on the size and sparsity of the derived graph.

In this sense, the augmented Hasse graph does not merely reduce topological representation learning to graph representation learning at a single rank; rather, it reduces the \emph{full multi-rank problem} to a single graph embedding problem on a structured heterogeneous graph, in which heterogeneity of node type---encoded by rank labels---is reflected in the edge structure of $\mathcal{H}_{\mathcal{X}}(\mathsf{N})$ and, consequently, in the geometry of the learned embedding space.
The algorithms described in \cref{section:topoembedx-overview} use this reduction at two levels of granularity.
Some methods first select a rank and a neighborhood relation, thereby embedding the corresponding same-rank graph induced by the augmented Hasse construction.
Other methods operate directly on an augmented Hasse graph containing cells from several ranks and therefore learn joint multi-rank embeddings.
All cases use graph representation learning as the computational backend; they differ in whether the learned representation is rank-restricted or multi-rank at the cell level.

\section{Overview of \texttt{TopoEmbedX}}%
\label{section:topoembedx-overview}

\texttt{TopoEmbedX}\footnote{\texttt{TopoEmbedX} is available at \url{https://github.com/pyt-team/topoembedx/}.} provides a software framework for higher-order representation learning across \emph{simplicial complexes}, \emph{hypergraphs}, \emph{cell complexes}, and \emph{CCs}.
The core computational principle underlying \texttt{TopoEmbedX} is the theoretical reduction of higher-order representation learning to graph representation learning, as described in the previous section.
To achieve this, \texttt{TopoEmbedX} transforms a given higher-order domain into its corresponding augmented Hasse graph and then applies established graph representation learning algorithms to compute embeddings for its cells.
The structural correspondence between the selected augmented Hasse subgraph and the original higher-order domain then directly yields well-defined embeddings for the cells represented in that subgraph.
Since the augmented Hasse graph is directed, symmetrization is performed at the embedding stage when an undirected input is required, by replacing each directed edge with an undirected edge.
We write $\mathcal{H}^{\mathrm{sym}}_{\mathcal{X}}(\mathsf{N})$ for this symmetrized augmented Hasse graph.
\texttt{TopoEmbedX} exposes a unified API that provides a consistent, interoperable interface across all topological domains supported by \texttt{TopoNetX}, promoting modularity, extensibility, and reproducibility in higher-order machine learning research.

\texttt{TopoEmbedX} extends the \texttt{karateclub} library \citep{Rozemberczki:2020}, which provides analogous embedding functionality for graphs, and works in tandem with it to offer powerful tools for embedding topological data while preserving the structural properties of the underlying topological domains.
The algorithms implemented in \texttt{TopoEmbedX} can be grouped into two complementary families.
Rank-restricted algorithms, such as \texttt{DeepCell}, \texttt{Cell2Vec}, \texttt{CellDiff2Vec}, \texttt{HOLE}, \texttt{HOGLEE}, \texttt{ComplexNetMF}, and \texttt{ComplexRep}, construct a graph from a chosen rank-$r$ (co)adjacency relation and return embeddings for the selected cells.
Multi-rank algorithms, such as \texttt{ComplexRandNE}, \texttt{ComplexWalklets}, and \texttt{ComplexHeat}, use $\mathcal{H}^{\mathrm{sym}}_{\mathcal{X}}(\mathsf{N})$ and learn embeddings for cells across ranks in a shared representation space.
The following method sketches summarize the primary embedding algorithms currently exposed by \texttt{TopoEmbedX}.

\subsection{\texttt{Cell2Vec}}

\texttt{Cell2Vec} generalizes \texttt{Node2Vec}~\citep{Grover:2016} to topological domains $\mathcal{X}$ by constructing adjacency and co-adjacency matrices that encode higher-order relationships in topological complexes, rather than relying solely on standard graph adjacency matrices.
In this representation, $\mat{A}_{ij} = 1$ indicates that the elements $i$ and $j$ are related, thereby capturing interactions among edges, triangles, and higher-dimensional cells.
The resulting matrices are subsequently transformed into a graph $G = (\mathcal{V}, \mathcal{E})$, which preserves the topology of the underlying complex while providing a graph structure on which embeddings can be learned.
Random walks are then performed on $G$, with transition probabilities governed by the node2vec parameters $p$ and $q$: the return parameter $p$ controls the likelihood of immediately revisiting the previous node, while the in--out parameter $q$ biases walks toward more local, breadth-first exploration or more outward, depth-first exploration.

\begin{algorithm}[H]
    \caption{\texttt{Cell2Vec}: Learning Complex Embeddings via Biased Random Walks}
    \begin{algorithmic}[1]
        \REQUIRE Topological domain $\mathcal{X}$, target rank $r$, return parameter $p$, exploration parameter $q$, walk length $\ell$, walks per source cell $m$, embedding dimension $d$
        \ENSURE Embedding vectors $\{ \vec{z}_i \}_{i \in \mathcal{V}}$

        \STATE $\mat{A} \gets$ rank-$r$ (co)adjacency matrix of $\mathcal{X}$
        \STATE $G \gets$ graph induced by $\mat{A}$
        \STATE $\mathcal{W} \gets$ $m$ length-$\ell$ biased walks from each $v\in\mathcal{V}$ using transition parameters $p,q$
        \STATE Fit skip-gram embeddings on walk-context pairs from $\mathcal{W}$: \\
        $\displaystyle \max \sum_{(i,j) \in \operatorname{pairs}(\mathcal{W})} \log \frac{\exp(\vec{z}_i^\top \vec{z}_j)}{\sum_{k \in \mathcal{V}} \exp(\vec{z}_i^\top \vec{z}_k)}$
        \RETURN $\{\vec{z}_i\}_{i \in \mathcal{V}}$
    \end{algorithmic}
\end{algorithm}

\subsection{\texttt{DeepCell}}
\texttt{DeepCell} generalizes \texttt{DeepWalk}~\citep{Perozzi:2014} to $\mathcal{X}$ by performing random walks over the incidence
structure of a complex, guided by adjacency and co-adjacency relations as well as incidence relations between cells of different ranks.
These walks generate sequences of cells, which are then used to train a skip-gram model similar to DeepWalk.
The key advantage of \texttt{DeepCell} is its flexibility: it allows users to define what ``neighborhood'' means in a topological domain and to learn embeddings that reflect both pairwise connections and higher-order topological relationships, making it a powerful unified approach to representation learning.

\begin{algorithm}[H]
    \caption{\texttt{DeepCell}: DeepWalk-based Embedding of $k$-Cells}

    \begin{algorithmic}[1]
        \REQUIRE Topological domain $\mathcal{X}$, target rank $k$, neighborhood collection $\mathsf{N}$,
        walk number $r$, walk length $L$, window size $w$, embedding dim $d$
        \ENSURE Embedding vectors $\{\vec{h}_c\}_{c \in \mathcal{X}^k}$

        \STATE $G_{\mathsf{N}} \gets$ cell graph whose nodes are selected cells and whose edges are induced by adjacency, co-adjacency, and incidence relations in $\mathsf{N}$; add self-loops
        \STATE $\mathcal{W} \gets$ $r$ length-$L$ random walks from each cell in $G_{\mathsf{N}}$
        \STATE Train skip-gram on $\mathcal{W}$ with window $w$ to obtain $\mathbf{h}_c \in \mathbb{R}^d$ for each cell $c$ in $G_{\mathsf{N}}$
        \RETURN $\{\vec{h}_c\}_{c \in \mathcal{X}^k}$
    \end{algorithmic}
\end{algorithm}

\subsection{\texttt{CellDiff2Vec}}
\texttt{CellDiff2Vec} is a diffusion-based topological embedding algorithm that generalizes \texttt{Diff2Vec}~\citep{Rozemberczki:2018} to $\mathcal{X}$.
In contrast to random walk-based algorithms, \texttt{CellDiff2Vec} simulates diffusion processes on a graph $G = (\mathcal{V}, \mathcal{E})$ built from the (co)adjacency relations encoded in the matrix $\mat{A}$, which is constructed from $\mathcal{X}$ using the specified \texttt{matrix\_type} and \texttt{rank}.
These diffusions are initiated from each cell $v \in \mathcal{V}$ and repeated $L$ times, producing sequences $S_v^i$ of length $C$, which capture the flow of information or connectivity through the topological structure.
By training a skip-gram model on the full set of diffusion-generated sequences $\{ S_v^i \}$, \texttt{CellDiff2Vec} learns $d$-dimensional embeddings $\{\mathbf{z}_v \in \mathbb{R}^d\}_{v \in \mathcal{V}}$ that encode both local and global relationships among cells, reflecting the underlying topology more robustly than simple random walks.
\begin{algorithm}[H]
    \caption{\texttt{CellDiff2Vec}: Diffusion-based Embeddings for Cells in Complexes}
    \label{alg:celldiff2vec}
    \begin{algorithmic}[1]
        \REQUIRE Topological domain $\mathcal{X}$, matrix selector $\tau$, target rank $r$, diffusion count $L$, diffusion length $C$, embedding dimension $d$
        \ENSURE Embedding $\vec{z}_v$ for each $v \in \mathcal{X}_r$

        \STATE $\mat{A} \gets$ rank-$r$ (co)adjacency matrix of $\mathcal{X}$ selected by $\tau$
        \STATE $G \gets$ graph induced by $\mat{A}$ on $\mathcal{X}_r$
        \STATE $\mathcal{S} \gets \{S_v^i: v\in\mathcal{V},\, i=1,\ldots,L\}$, where $S_v^i$ is a length-$C$ diffusion from $v$ on $G$
        \STATE Fit a skip-gram model on $\mathcal{S}$ to obtain $d$-dimensional cell embeddings
        \RETURN $\{ \vec{z}_v \in \mathbb{R}^d \}_{v \in \mathcal{V}}$
    \end{algorithmic}
\end{algorithm}
\subsection{\texttt{HOLE}}
Higher Order Laplacian Eigenmaps (\texttt{HOLE}) is a spectral embedding algorithm that generalizes \texttt{Laplacian Eigenmaps}~\citep{Belkin:2001} to $\mathcal{X}$.
The core idea is to capture the geometry of the complex by representing its cells and their (co)adjacency relations as a graph, and then using the spectrum of the graph Laplacian to obtain embeddings.
By computing the eigenvectors associated with the smallest nonzero eigenvalues of the Laplacian, the algorithm produces embeddings that preserve local neighborhood information and reflect the intrinsic geometry of the data.

\begin{algorithm}[H]
    \caption{Higher Order Laplacian Eigenmaps}\label{alg:ho_laplacian_eigenmaps}
    \begin{algorithmic}[1]
        \REQUIRE Topological domain $\mathcal{X}$, matrix selector $\tau$, target rank $r$, embedding dimension $d$

        \STATE $\mat{A} \gets$ rank-$r$ (co)adjacency matrix selected by $\tau$
        \STATE $G \gets$ graph induced by $\mat{A}$ on $\mathcal{X}_r$
        \STATE $\{ \vec{u}_i\}_{i=1}^d \gets$ eigenvectors associated with the smallest nonzero eigenvalues of the normalized graph Laplacian of $G$
        \STATE $\vec{z}_v \gets (\vec{u}_1(v), \ldots, \vec{u}_d(v))$ for each $v \in \mathcal{V}_G$
        \RETURN $\{ \vec{z}_v \in \mathbb{R}^d \}_{v \in \mathcal{V}_G}$
    \end{algorithmic}
\end{algorithm}

\subsection{\texttt{HOGLEE}}
Higher Order Geometric Laplacian Eigenmaps (\texttt{HOGLEE}) is a geometric spectral embedding algorithm designed to capture the rich structure of higher-order complexes, such as CCs.
It generalizes \texttt{Geometric Laplacian Eigenmaps}~\citep{Torres:2020} to this setting.
\texttt{HOGLEE} incorporates geometric information about the relationships between cells, including features such as shortest paths and commute times, into a geometric Laplacian matrix together with the adjacency.
The algorithm constructs a graph representation of the complex, computes relevant geometric features, and then forms a Laplacian that encodes both topological and geometric properties.
By finding the eigenvectors associated with the smallest nonzero eigenvalues of this geometric Laplacian, \texttt{HOGLEE} produces embeddings that reflect both the local and global geometry of the complex.

\begin{algorithm}[H]
    \caption{\texttt{HOGLEE}: Higher Order Geometric Laplacian Eigenmaps}
    \label{alg:hoglee}
    \begin{algorithmic}[1]
        \REQUIRE Topological domain $\mathcal{X}$, matrix selector $\tau$, target rank $r$, embedding dimension $d$

        \STATE $\mat{A} \gets$ rank-$r$ (co)adjacency matrix selected by $\tau$
        \STATE $G \gets$ graph induced by $\mat{A}$ on $\mathcal{X}_r$
        \STATE $\Phi \gets$ geometric features of $G$ (e.g., shortest paths, commute times)
        \STATE $\mat{L}_g \gets$ geometric Laplacian induced by $\Phi$
        \STATE $\{ \vec{u}_i \}_{i=1}^d \gets$ eigenvectors of $\mat{L}_g$ associated with its smallest nonzero eigenvalues
        \STATE $\vec{z}_v \gets (\vec{u}_1(v), \ldots, \vec{u}_d(v))$ for each $v \in \mathcal{V}_G$
        \RETURN $\{ \vec{z}_v \in \mathbb{R}^d \}_{v \in \mathcal{V}_G}$
    \end{algorithmic}
\end{algorithm}

\subsection{\texttt{ComplexNetMF}}

\texttt{ComplexNetMF} generalizes \texttt{NetMF}~\citep{Qiu:2018} to $\mathcal{X}$ by creating a weighted graph that encodes the relationships between cells in $\mathcal{X}$.
These weights capture not only standard adjacency but also topological connectivity beyond edges, allowing the embeddings to reflect richer structural information.
The resulting weighted graph matrix is factorized via truncated \texttt{SVD}, producing embeddings that preserve both walk-based proximities and the higher-order topological structure of the data.
This topological extension makes \texttt{ComplexNetMF} well-suited for complex datasets where higher-order interactions drive the underlying geometry.
As in common implementations of \texttt{NetMF}, the matrix factorization is performed on the largest connected component of the induced graph.
Cells outside this component are not included in the factorization.

\begin{algorithm}[H]
    \caption{\texttt{ComplexNetMF}}
    \label{alg:ComplexNetMF}
    \begin{algorithmic}[1]
        \REQUIRE Topological domain $\mathcal{X}$, matrix selector $\tau$, target rank $r$, embedding dimension $d$, window size $w$, optional via-rank $s$

        \STATE $\mat{A} \gets$ rank-$r$ (co)adjacency matrix selected by $\tau$ and optional via-rank $s$
        \STATE $G \gets$ largest connected component of graph induced by $\mat{A}$ with self-loops
        \STATE $\mat{M} \gets$ NetMF/DeepWalk PMI matrix of $G$ with window size $w$ \hfill (cf. \citep{Qiu:2018})
        \STATE $\mat{U}_d\mat{\Sigma}_d\mat{V}_d^\top \gets$ rank-$d$ truncated SVD of $\mat{M}$; $\mat{Z}\gets\mat{U}_d\mat{\Sigma}_d$
        \RETURN $\{ \vec{z}_v \in \mathbb{R}^d \}_{v \in \mathcal{V}_G}$ together with the corresponding original cell identifiers
    \end{algorithmic}
\end{algorithm}

\subsection{\texttt{ComplexRep}}
\texttt{ComplexRep} generalizes \texttt{GraRep}~\citep{Cao:2015} to $\mathcal{X}$ by
replacing simple adjacency matrices with (co)adjacency matrices defined on topological
domains, thereby encoding relationships between higher-dimensional cells.
Each power of these topological (co)adjacency matrices reflects increasingly global
structure, and embeddings are then obtained through matrix factorization.
By embedding complexes in this way, \texttt{ComplexRep} produces representations that
incorporate multi-scale topological interactions.

\begin{algorithm}[H]
    \caption{\texttt{ComplexRep}}
    \label{alg:ComplexRep}
    \begin{algorithmic}[1]
        \REQUIRE Topological domain $\mathcal{X}$, matrix selector $\tau$, target
        rank $r$, embedding dimension $d$, scale count $K$, optional via-rank $s$
        \STATE $\tilde{\mat{A}} \gets$ rank-$r$ (co)adjacency selected by $\tau$
        and optional via-rank $s$, with self-loops added
        \STATE $\mat{T} \gets \tilde{\mat{D}}^{-1} \tilde{\mat{A}}$, where
        $\tilde{\mat{D}}$ is the diagonal degree matrix of $\tilde{\mat{A}}$
        \STATE Set $d_k = \lfloor d/K \rfloor + \mathbf{1}[k \leq d \bmod K]$
        for $k = 1, \dots, K$, so that $\sum_{k=1}^{K} d_k = d$
        \FOR{$k = 1, \dots, K$}
        \STATE $\mat{M}^{(k)} \gets$ GraRep matrix derived from $\mat{T}^k$
        \hfill (e.g., log/PPMI transform, cf.\ \citep{Cao:2015})
        \STATE $\mat{U}_{d_k}^{(k)} \mat{\Sigma}_{d_k}^{(k)}
        {\mat{V}_{d_k}^{(k)}}^\top \gets$ rank-$d_k$ truncated SVD of
        $\mat{M}^{(k)}$
        \STATE $\mat{Z}^{(k)} \gets \mat{U}_{d_k}^{(k)} \mat{\Sigma}_{d_k}^{(k)}$
        \ENDFOR
        \STATE $\mat{Z} \gets [\, \mat{Z}^{(1)} \; \| \; \cdots \; \| \;
        \mat{Z}^{(K)} \,]$
        \RETURN $\{ \vec{z}_v \in \mathbb{R}^d \}_{v \in \mathcal{V}}$,
        preserving original cell ordering
    \end{algorithmic}
\end{algorithm}

\subsection{\texttt{ComplexRandNE}}
\texttt{ComplexRandNE} generalizes \texttt{RandNE}~\citep{Zhang:2018} to $\mathcal{X}$ by applying randomized linear projections to a matrix representation derived from $\mathcal{H}^{\mathrm{sym}}_{\mathcal{X}}(\mathsf{N})$. This yields low-dimensional embeddings for cells across multiple ranks in a small number of sparse matrix multiplications, making the method computationally efficient and scalable for large topological domains.

\begin{algorithm}[H]
    \caption{\texttt{ComplexRandNE}: Randomized Projection Embeddings on the Augmented Hasse Graph}
    \label{alg:ComplexRandNE}
    \begin{algorithmic}[1]
        \REQUIRE Topological domain $\mathcal{X}$, neighborhood collection $\mathsf{N}$, embedding dimension $d$, propagation depth $K$, smoothing weights $\{\alpha_k\}_{k=0}^K$
        \ENSURE Embeddings $\{ \vec{z}_x \in \mathbb{R}^d \}_{x \in \mathcal{X}}$

        \STATE $\mat{S} \gets$ adjacency matrix of $\mathcal{H}^{\mathrm{sym}}_{\mathcal{X}}(\mathsf{N})$
        \STATE $\mat{Z}^{(0)} \gets$ random matrix in $\mathbb{R}^{|\mathcal{X}| \times d}$
        \FOR{$k = 1$ to $K$}
        \STATE $\mat{Z}^{(k)} \gets \mat{S} \mat{Z}^{(k-1)}$
        \ENDFOR
        \STATE $\mat{Z} \gets \sum_{k=0}^{K} \alpha_k \mat{Z}^{(k)}$; normalize rows of $\mat{Z}$
        \RETURN $\{ \vec{z}_x \}_{x \in \mathcal{X}}$
    \end{algorithmic}
\end{algorithm}

\subsection{\texttt{ComplexWalklets}}

\texttt{ComplexWalklets} is a higher-order generalization of the \texttt{Walklets}~\citep{Perozzi:2017} algorithm for $\mathcal{X}$.
The method performs dimension-aware random walks on $\mathcal{H}^{\mathrm{sym}}_{\mathcal{X}}(\mathsf{N})$ and applies skip-step sampling to capture multi-scale incidence relationships among cells.
The resulting sequences are used to train a skip-gram model that yields topology-preserving embeddings for cells across multiple dimensions.
\begin{algorithm}[H]
    \caption{\texttt{ComplexWalklets}: Multi-scale Skip-step Walks on the Augmented Hasse Graph}
    \label{alg:ComplexWalklets}
    \begin{algorithmic}[1]
        \REQUIRE Topological domain $\mathcal{X}$, neighborhood collection $\mathsf{N}$, walk length $L$, walks per cell $R$, maximum skip $K$, embedding dimension $d$
        \ENSURE Embeddings $\{ \vec{z}_x \in \mathbb{R}^d \}_{x \in \mathcal{X}}$

        \STATE $\mathcal{W}\gets$ $R$ length-$L$ walks from each $x\in\mathcal{X}$ on $\mathcal{H}^{\mathrm{sym}}_{\mathcal{X}}(\mathsf{N})$
        \STATE $\mathcal{C}\gets\{\operatorname{skip}_k(W): W\in\mathcal{W},\, k=1,\ldots,K\}$
        \STATE Fit a skip-gram model on $\mathcal{C}$
        \RETURN $\{ \vec{z}_x \}_{x \in \mathcal{X}}$
    \end{algorithmic}
\end{algorithm}

\subsection{\texttt{ComplexHeat}}

\texttt{ComplexHeat} extends heat-kernel graph embeddings \citep{Tsitsulin:2018} to $\mathcal{X}$ by defining a heat diffusion operator on $\mathcal{H}^{\mathrm{sym}}_{\mathcal{X}}(\mathsf{N})$.
The method models heat propagation across cells of varying dimensions and employs the resulting diffusion signatures as embeddings, thereby capturing multi-scale topological structure and diffusion geometry within the complex.

\begin{algorithm}[H]
    \caption{\texttt{ComplexHeat}: Heat-kernel Embeddings on the Augmented Hasse Graph}
    \label{alg:ComplexHeat}
    \begin{algorithmic}[1]
        \REQUIRE Topological domain $\mathcal{X}$, neighborhood collection $\mathsf{N}$, diffusion times $\{t_1,\dots,t_m\}$, embedding dimension $d$
        \ENSURE Embeddings $\{ \vec{z}_x \in \mathbb{R}^d \}_{x \in \mathcal{X}}$

        \STATE $\mat{L} \gets$ normalized Laplacian of $\mathcal{H}^{\mathrm{sym}}_{\mathcal{X}}(\mathsf{N})$
        \STATE $\mat{F}\gets [\,\mat{F}_{t_1}\|\cdots\|\mat{F}_{t_m}\,]$, where $\mat{F}_{t_j}$ contains row-wise signatures of $\exp(-t_j\mat{L})$
        \STATE If needed, reduce $\mat{F}$ to dimension $d$ via truncated SVD or PCA
        \RETURN $\{ \vec{z}_x \}_{x \in \mathcal{X}}$
    \end{algorithmic}
\end{algorithm}

\section{Experimental Setup}%
\label{section:experiments}

\begin{figure}
    \resizebox{\textwidth}{!}{%
\begin{tikzpicture}[
        edge baseline/.style={bar width=3pt, bar shift=-15pt, fill=rwth-blue-75, draw=black!70, postaction={pattern=horizontal lines, pattern color=black}},
        edge celltwo/.style={bar width=3pt, bar shift=-12pt, fill=rwth-maygreen-75, draw=black!70, postaction={pattern=vertical lines, pattern color=black}},
        edge celldiff/.style={bar width=3pt, bar shift=-9pt, fill=rwth-orange-75, draw=black!70, postaction={pattern=north east lines, pattern color=black}},
        edge complexheat/.style={bar width=3pt, bar shift=-6pt, fill=rwth-green-75, draw=black!70, postaction={pattern=dots, pattern color=black}},
        edge complexnetmf/.style={bar width=3pt, bar shift=-3pt, fill=rwth-purple-75, draw=black!70, postaction={pattern=crosshatch dots, pattern color=black}},
        edge complexrandne/.style={bar width=3pt, bar shift=0pt, fill=rwth-turquoise-75, draw=black!70, postaction={pattern=bricks, pattern color=black}},
        edge complexrep/.style={bar width=3pt, bar shift=3pt, fill=rwth-magenta-75, draw=black!70, postaction={pattern=fivepointed stars, pattern color=black}},
        edge complexwalklets/.style={bar width=3pt, bar shift=6pt, fill=rwth-bordeaux-75, draw=black!70, postaction={pattern=checkerboard, pattern color=black}},
        edge deepcell/.style={bar width=3pt, bar shift=9pt, fill=rwth-petrol-75, draw=black!70, postaction={pattern=north west lines, pattern color=black}},
        edge hoglee/.style={bar width=3pt, bar shift=12pt, fill=rwth-violet-75, draw=black!70, postaction={pattern=grid, pattern color=black}},
        edge hole/.style={bar width=3pt, bar shift=15pt, fill=rwth-red-75, draw=black!70, postaction={pattern=crosshatch, pattern color=black}}
    ]
    \begin{groupplot}[
            group style={group size=2 by 1, horizontal sep=1.2cm},
            height=4.2cm,
            ybar,
            ymajorgrids,
            scale only axis,
            x tick label style={anchor=north,font=\scriptsize,align=center},
            tick pos=left,
        ]

        \nextgroupplot[
            width=10.1cm,
            ymin=0,
            ymax=1,
            ylabel={Accuracy},
            enlarge x limits=0.12,
            symbolic x coords={
                algebra-questions,
                cooking,
                geometry-questions,
                madison-restaurant-reviews,
                MAG-10,
                music-blues-reviews,
                vegas-bars-reviews
            },
            xtick={
                algebra-questions,
                cooking,
                geometry-questions,
                madison-restaurant-reviews,
                MAG-10,
                music-blues-reviews,
                vegas-bars-reviews
            },
            xticklabels={
                \shortstack{algebra\\questions},
                cooking,
                \shortstack{geometry\\questions},
                \shortstack{madison\\restaurant\\reviews},
                MAG-10,
                \shortstack{music-blues\\reviews},
                \shortstack{vegas-bars\\reviews}
            },
            legend to name=edge-task-legend,
            legend columns=6,
            legend cell align=left,
            legend style={
                draw=none,
                font=\scriptsize,
                fill=none,
                /tikz/every even column/.append style={column sep=10pt}
            }
        ]

        \addplot+[edge baseline] table[
            col sep=comma,
            x=Dataset,
            y=Accuracy,
            discard if not={Algorithm}{baseline_majority}
        ] {figures/edge_classification.csv};
        \addlegendentry{baseline}

        \addplot+[edge celltwo] table[
            col sep=comma,
            x=Dataset,
            y=Accuracy,
            discard if not={Algorithm}{cell2vec}
        ] {figures/edge_classification.csv};
        \addlegendentry{\texttt{Cell2Vec}}

        \addplot+[edge celldiff] table[
            col sep=comma,
            x=Dataset,
            y=Accuracy,
            discard if not={Algorithm}{celldiff2vec}
        ] {figures/edge_classification.csv};
        \addlegendentry{CellDiff2Cec}

        \addplot+[edge complexheat] table[
            col sep=comma,
            x=Dataset,
            y=Accuracy,
            discard if not={Algorithm}{ComplexHeat}
        ] {figures/edge_classification.csv};
        \addlegendentry{\texttt{ComplexHeat}}

        \addplot+[edge complexnetmf] table[
            col sep=comma,
            x=Dataset,
            y=Accuracy,
            discard if not={Algorithm}{ComplexNetMF}
        ] {figures/edge_classification.csv};
        \addlegendentry{\texttt{ComplexNetMF}}

        \addplot+[edge complexrandne] table[
            col sep=comma,
            x=Dataset,
            y=Accuracy,
            discard if not={Algorithm}{ComplexRandNE}
        ] {figures/edge_classification.csv};
        \addlegendentry{\texttt{ComplexRandNE}}

        \addlegendimage{ybar, edge complexrep}
        \addlegendentry{\texttt{ComplexRep}}

        \addplot+[edge complexwalklets] table[
            col sep=comma,
            x=Dataset,
            y=Accuracy,
            discard if not={Algorithm}{ComplexWalklets}
        ] {figures/edge_classification.csv};
        \addlegendentry{\texttt{ComplexWalklets}}

        \addplot+[edge deepcell] table[
            col sep=comma,
            x=Dataset,
            y=Accuracy,
            discard if not={Algorithm}{deepcell}
        ] {figures/edge_classification.csv};
        \addlegendentry{\texttt{DeepCell}}

        \addplot+[edge hoglee] table[
            col sep=comma,
            x=Dataset,
            y=Accuracy,
            discard if not={Algorithm}{hoglee}
        ] {figures/edge_classification.csv};
        \addlegendentry{\texttt{HOGLEE}}

        \addplot+[edge hole] table[
            col sep=comma,
            x=Dataset,
            y=Accuracy,
            discard if not={Algorithm}{hole}
        ] {figures/edge_classification.csv};
        \addlegendentry{\texttt{HOLE}}

        \nextgroupplot[
            width=1.6cm,
            ymin=0,
            ylabel={MSE},
            xmin=-0.5,
            xmax=0.5,
            x tick label style={anchor=north,font=\scriptsize,align=center},
            xtick={0},
            xticklabels={
                \shortstack{semantic-scholar\\coauth-sample}
            }
        ]

        \addplot+[edge baseline] table[
            col sep=comma,
            x expr=0,
            y=MSE,
            discard if not={Algorithm}{baseline_mean}
        ] {figures/edge_regression.csv};

        \addplot+[edge celltwo] table[
            col sep=comma,
            x expr=0,
            y=MSE,
            discard if not={Algorithm}{cell2vec}
        ] {figures/edge_regression.csv};

        \addplot+[edge celldiff] table[
            col sep=comma,
            x expr=0,
            y=MSE,
            discard if not={Algorithm}{celldiff2vec}
        ] {figures/edge_regression.csv};

        \addplot+[edge complexheat] table[
            col sep=comma,
            x expr=0,
            y=MSE,
            discard if not={Algorithm}{ComplexHeat}
        ] {figures/edge_regression.csv};

        \addplot+[edge complexnetmf] table[
            col sep=comma,
            x expr=0,
            y=MSE,
            discard if not={Algorithm}{ComplexNetMF}
        ] {figures/edge_regression.csv};

        \addplot+[edge complexrandne] table[
            col sep=comma,
            x expr=0,
            y=MSE,
            discard if not={Algorithm}{ComplexRandNE}
        ] {figures/edge_regression.csv};

        \addplot+[edge complexrep] table[
            col sep=comma,
            x expr=0,
            y=MSE,
            discard if not={Algorithm}{ComplexRep}
        ] {figures/edge_regression.csv};

        \addplot+[edge complexwalklets] table[
            col sep=comma,
            x expr=0,
            y=MSE,
            discard if not={Algorithm}{ComplexWalklets}
        ] {figures/edge_regression.csv};

        \addplot+[edge deepcell] table[
            col sep=comma,
            x expr=0,
            y=MSE,
            discard if not={Algorithm}{deepcell}
        ] {figures/edge_regression.csv};

        \addplot+[edge hoglee] table[
            col sep=comma,
            x expr=0,
            y=MSE,
            discard if not={Algorithm}{hoglee}
        ] {figures/edge_regression.csv};

        \addplot+[edge hole] table[
            col sep=comma,
            x expr=0,
            y=MSE,
            discard if not={Algorithm}{hole}
        ] {figures/edge_regression.csv};
    \end{groupplot}

    \node[anchor=north] at (current bounding box.south) {\pgfplotslegendfromname{edge-task-legend}};
\end{tikzpicture}
    }
    \caption{%
        Edge classification results across different datasets (left) and edge regression results on \texttt{semantic-scholar-coauth-sample} (right). Missing values indicate algorithm-dataset combinations that reached the one-hour timeout described in the text.
    }
    \label{figure:edge-tasks}
\end{figure}
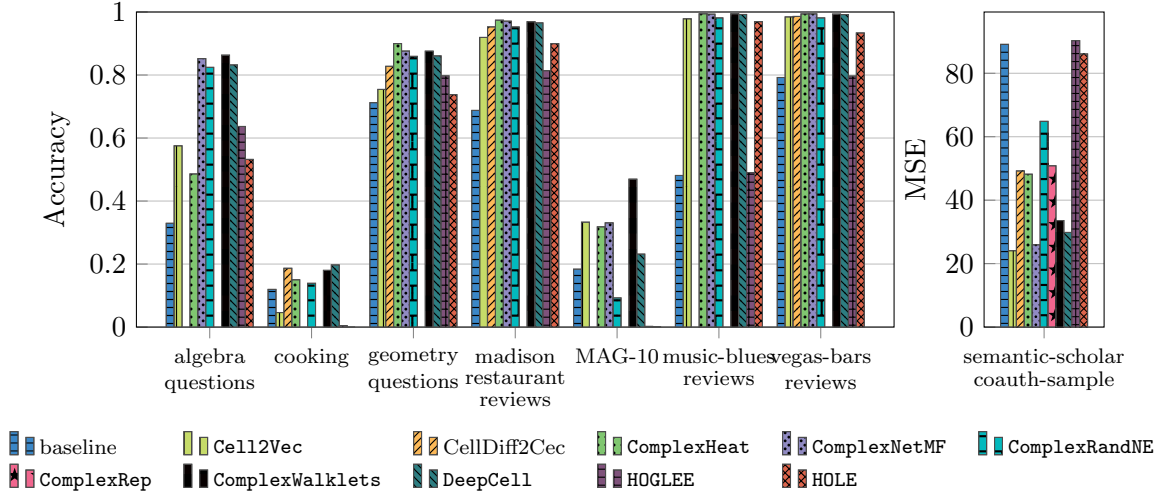

We evaluate the algorithms in \texttt{TopoEmbedX} on two transductive supervised edge-level tasks: edge classification and edge regression.
The datasets used in these experiments were obtained from AHORN, an online collection of higher-order datasets\footnote{\url{https://ahorn.rwth-aachen.de}}.
In this transductive setting, the full simplicial complex is available when computing unsupervised cell embeddings, while the edge labels or regression targets are split into fixed 80/20 train and test sets for downstream supervised evaluation.
Algorithm-specific hyperparameters were left at their default values (corresponding to the analogous parameters in \texttt{karateclub}; see also Appendix~\ref{appendix:hyperparameters}).
For edge classification, we convert each dataset into a simplicial complex, extract labels from edge metadata, compute rank-1 embeddings from co-adjacency neighborhoods via 2-cells, and train an MLP classifier.
The left panel of \cref{figure:edge-tasks} reports classification accuracy against a majority-class baseline on the seven datasets \texttt{algebra-questions}, \texttt{cooking}, \texttt{geometry-questions}, \texttt{madison-restaurant-reviews}, \texttt{MAG-10}, \texttt{music-blues-reviews}, and \texttt{vegas-bars-reviews}.
For edge regression, we use the same overall recipe on \texttt{semantic-scholar-coauth-sample}, predict the scalar edge attribute with an MLP regressor using the same optimization settings, and report mean squared error (MSE) relative to a simple mean baseline in the right panel of \cref{figure:edge-tasks}.
The classification baseline always predicts the most frequent class in the training split, while the regression baseline predicts the training-set mean of the target attribute.
Exact results are provided in the appendix.

\Cref{figure:edge-tasks} shows a clear pattern in the current results.
In edge classification, \texttt{ComplexWalklets} attains the best reported accuracies on \texttt{algebra-questions}, \texttt{MAG-10}, and \texttt{music-blues-reviews}.
\texttt{ComplexHeat} remains strongest on \texttt{geometry-questions} and \texttt{madison-restaurant-reviews}, and is nearly tied with \texttt{ComplexWalklets} on \texttt{music-blues-reviews}.
\texttt{ComplexNetMF} achieves the top score on \texttt{vegas-bars-reviews} and remains competitive on several other datasets.
The main remaining exception is the more difficult \texttt{cooking} setting, where \texttt{DeepCell} achieves the top score.
\texttt{DeepCell} is otherwise typically the closest previously published method, while \texttt{ComplexRandNE} remains competitive on several datasets but degrades substantially on \texttt{MAG-10}.
The walk-based algorithms \texttt{Cell2Vec}, \texttt{CellDiff2Vec}, and \texttt{ComplexWalklets} are also strong where available, especially on the review and question datasets, but several method-dataset pairs are missing because they reached the evaluation timeout.
The most difficult classification settings are \texttt{MAG-10} and \texttt{cooking}: on \texttt{MAG-10}, all algorithms remain far below the near-perfect accuracies observed on several other datasets, while on \texttt{cooking} even the best-performing algorithms achieve only modest improvements over the baseline.
In edge regression, \texttt{Cell2Vec} achieves the lowest MSE (24.07), closely followed by \texttt{ComplexNetMF} (25.93), \texttt{DeepCell} (29.83), and \texttt{ComplexWalklets} (33.58), all of which substantially outperform the mean baseline (89.11).
\texttt{ComplexHeat} (48.19) performs similarly to \texttt{CellDiff2Vec} (49.21) and \texttt{ComplexRep} (50.84), while \texttt{ComplexRandNE} provides a more modest gain.
By contrast, \texttt{HOGLEE} and \texttt{HOLE} do not materially improve on the baseline in this experiment.

For these experiments, each algorithm-dataset run was stopped after a one-hour wall-clock timeout on an AMD EPYC 7763 CPU; no explicit memory limit was imposed.
All algorithms are implemented within the same software framework, which provides a common data interface and reduces implementation-level confounders. Source code is available\footnote{\url{https://github.com/pyt-team/topoembedx-paper}} to support reproducibility.
Taken together, the experiments show that higher-order embeddings can provide strong edge-level predictors, but their effectiveness is dataset-dependent and their computational cost remains an important practical constraint.

\section{Conclusion}%
\label{section:conclusion}

\texttt{TopoEmbedX} provides a software layer for learning representations of topological domains in a unified manner. By exposing a common interface for simplicial complexes, hypergraphs, cell complexes, and CCs, the library makes it possible to construct neighborhood-based cell graphs, augmented Hasse graphs, and rank-restricted subgraphs through a unified workflow. This design allows researchers to apply, compare, and extend a broad family of embedding algorithms without rewriting the data-handling and graph-construction machinery for each topological domain or each target rank.

The experiments demonstrate that these embeddings can serve as useful features for supervised edge-level prediction tasks.
Across the datasets considered here, several topological embedding techniques substantially outperform the baselines, although the strongest method depends on the dataset and task.
The results also show that no single algorithm dominates uniformly: walk-based, factorization-based, and diffusion-based approaches each perform well in different settings.
These findings support the value of a unified benchmarking framework in which such methodological differences can be assessed under a common data interface and evaluation protocol.

Several limitations remain.
The present evaluation focuses on edge classification and edge regression, and should be extended to additional tasks such as cell clustering and link prediction.
The experiments also rely mostly on default hyperparameters, so a fuller empirical study should include systematic model selection, repeated train-test splits, runtime and memory measurements, and larger datasets.
On the methodological side, the augmented Hasse graph reduction gives a flexible way to reuse graph embedding techniques, but further work is needed to understand when particular neighborhood choices preserve task-relevant topological information and when they introduce unnecessary computational cost.
Future versions of \texttt{TopoEmbedX} will therefore benefit from deeper theoretical analysis, broader benchmarks, and task-adaptive embedding techniques built on the same software foundation.

\section*{Acknowledgments}
F.~F.\ and M.~T.~S.\ acknowledge funding by the Ministry of Culture and Science (MKW) of the German State of North Rhine-Westphalia~(``NRW Rückkehrprogramm'') and the European Union (ERC, HIGH-HOPeS, 101039827).
Views and opinions expressed are, however, those of the authors only and do not necessarily reflect those of the European Union or the European Research Council Executive Agency.
Neither the European Union nor the granting authority can be held responsible for them.
M.~H.\ was supported in part by the National Science Foundation (NSF, DMS-2134231).

\bibliographystyle{plainnat}
\bibliography{references}

\appendix

\section{Dataset Statistics}%
\label{appendix:dataset-statistics}

The following table reports the structural and label statistics for the AHORN datasets used in the experiments.

\begin{table}[H]
    \centering
    \caption{Structural and label statistics for the datasets used in the edge-level experiments.}%
    \label{table:dataset-statistics}
    \begin{tabular}{lrrrr}
        \toprule
        Dataset                                 & Vertices    & Maximal Faces & Classes  & Imbalance Degree \\
        \midrule
        \texttt{algebra-questions}              & \num{423}   & \num{1268}    & \num{32} & \num{20.28} \\
        \texttt{cooking}                        & \num{6714}  & \num{39774}   & \num{20} & \num{13.31} \\
        \texttt{geometry-questions}             & \num{580}   & \num{1193}    & \num{25} & \num{15.28} \\
        \texttt{madison-restaurant-reviews}     & \num{565}   & \num{601}     & \num{9}  & \num{4.31} \\
        \texttt{MAG-10}                         & \num{80198} & \num{51889}   & \num{10} & \num{5.22} \\
        \texttt{music-blues-reviews}            & \num{1106}  & \num{694}     & \num{7}  & \num{3.56} \\
        \texttt{vegas-bars-reviews}             & \num{1234}  & \num{1194}    & \num{15} & \num{5.33} \\
        \texttt{semantic-scholar-coauth-sample} & \num{352}   & \num{24200}   & -        & - \\
        \bottomrule
    \end{tabular}
\end{table}

\section{Experimental Hyperparameters}%
\label{appendix:hyperparameters}

All embedding algorithms were executed using their default constructor hyperparameters.
For algorithms that serve as topological analogues of implementations in \texttt{karateclub}, the corresponding \texttt{karateclub} default settings were used where applicable.
\Cref{table:embedding-hyperparameters} summarizes the hyperparameters employed in these experiments.

\begin{table}[H]
    \centering
    \caption{Embedding hyperparameters used in the edge-level experiments.}%
    \label{table:embedding-hyperparameters}
    \begin{tabular}{p{0.15\textwidth}p{0.8\textwidth}}
        \toprule
        Algorithm                & Hyperparameters \\
        \midrule
        \texttt{Cell2Vec}        &
        \texttt{dimensions=128}, \texttt{walk\_number=10}, \texttt{walk\_length=80}, \texttt{p=1.0}, \texttt{q=1.0}, \texttt{window\_size=5} \\
        \texttt{DeepCell}        &
        \texttt{dimensions=128}, \texttt{walk\_number=10}, \texttt{walk\_length=80}, \texttt{window\_size=5} \\
        \texttt{CellDiff2Vec}    &
        \texttt{dimensions=128}, \texttt{diffusion\_number=10}, \texttt{diffusion\_cover=80}, \texttt{window\_size=5} \\
        \texttt{HOLE}            &
        \texttt{dimensions=3}, \texttt{maximum\_number\_of\_iterations=100} \\
        \texttt{HOGLEE}          &
        \texttt{dimensions=128} \\
        \texttt{ComplexHeat}     &
        \texttt{sample\_number=200}, \texttt{step\_size=0.1}, \texttt{heat\_coefficient=1.0}, \texttt{approximation=100}, \texttt{mechanism=approximate} \\
        \texttt{ComplexNetMF}    &
        \texttt{dimensions=32}, \texttt{iteration=10}, \texttt{order=2}, \texttt{negative\_samples=1} \\
        \texttt{ComplexRandNE}   &
        \texttt{dimensions=128}, \texttt{alphas=[0.5, 0.5]} \\
        \texttt{ComplexRep}      &
        \texttt{dimensions=32}, \texttt{iteration=10}, \texttt{order=5} \\
        \texttt{ComplexWalklets} &
        \texttt{dimensions=32}, \texttt{walk\_number=10}, \texttt{walk\_length=80}, \texttt{window\_size=4} \\
        \bottomrule
    \end{tabular}
\end{table}

The downstream supervised MLPs were implemented with \texttt{scikit-learn} \citep{Pedregosa:2011}.
For both the classifier and regressor, we used one hidden layer with \num{100} units, ReLU activations, the Adam solver, an $\ell_2$ regularization coefficient of \num{0.0001}, adaptive learning rates, an initial learning rate of \num{0.001}, at most \num{10000} iterations, and a convergence tolerance of \num{0.0001}.

\section{Experimental Results}%
\label{appendix:experimental-results}

\Cref{table:edge-classification-results,table:edge-regression-results} report the exact numerical results for the edge classification and edge regression experiments, respectively, that are summarized in \cref{figure:edge-tasks}.
\texttt{ComplexRep} is missing from the edge classification results because it reached the one-hour timeout for every dataset.

\pgfplotstableread[col sep=comma]{figures/edge_classification.csv}\edgeclassificationresults
\pgfplotstableread[col sep=comma]{figures/edge_regression.csv}\edgeregressionresults

\newcolumntype{C}[1]{>{\centering\arraybackslash}p{#1}}
\newcommand{\datasetheader}[1]{\texttt{\StrSubstitute{#1}{-}{-\allowbreak}}}

\newcommand{\csvresultlookup}[5]{%
    \def#5{}%
    \edef\resultdataset{#3}%
    \edef\resultalgorithm{#4}%
    \pgfplotstablegetrowsof{#1}%
    \pgfmathtruncatemacro{\lasttablerow}{\pgfplotsretval-1}%
    \pgfplotsforeachungrouped \tablerow in {0,...,\lasttablerow}{%
        \pgfplotstablegetelem{\tablerow}{Dataset}\of#1%
        \edef\currentdataset{\pgfplotsretval}%
        \ifnum\pdfstrcmp{\currentdataset}{\resultdataset}=0\relax%
        \pgfplotstablegetelem{\tablerow}{Algorithm}\of#1%
        \edef\currentalgorithm{\pgfplotsretval}%
        \ifnum\pdfstrcmp{\currentalgorithm}{\resultalgorithm}=0\relax%
        \pgfplotstablegetelem{\tablerow}{#2}\of#1%
        \edef#5{\pgfplotsretval}%
        \fi%
        \fi%
    }%
}

\newcommand{\setresultcellcontent}{%
    \ifx\resultvalue\empty%
    \def\formattedresult{}%
    \else%
    \pgfmathprintnumberto[fixed, fixed zerofill, precision=4]{\resultvalue}{\formattedresult}%
    \fi%
    \pgfkeyslet{/pgfplots/table/create col/next content}\formattedresult%
}

\newcommand{\setedgeclassificationresultcell}[1]{%
    \getthisrow{AlgorithmKey}{\resultalgorithm}%
    \csvresultlookup{\edgeclassificationresults}{Accuracy}{#1}{\resultalgorithm}{\resultvalue}%
    \setresultcellcontent%
}

\newcommand{\setedgeregressionresultcell}[1]{%
    \getthisrow{AlgorithmKey}{\resultalgorithm}%
    \csvresultlookup{\edgeregressionresults}{MSE}{#1}{\resultalgorithm}{\resultvalue}%
    \setresultcellcontent%
}

\pgfplotstableset{
    result table setup/.style={
        string type,
        every head row/.style={
            before row=\toprule,
            after row=\midrule
        },
        every last row/.style={
            after row=\bottomrule
        },
        columns/Algorithm/.style={column type=l},
        columns/algebra-questions/.style={column name={\datasetheader{algebra-questions}}, column type=C{1.3cm}},
        columns/cooking/.style={column name={\datasetheader{cooking}}, column type=C{1.3cm}},
        columns/geometry-questions/.style={column name={\datasetheader{geometry-questions}}, column type=C{1.3cm}},
        columns/madison-restaurant-reviews/.style={column name={\datasetheader{madison-restaurant-reviews}}, column type=C{1.3cm}},
        columns/MAG-10/.style={column name={\datasetheader{MAG-10}}, column type=C{1.3cm}},
        columns/music-blues-reviews/.style={column name={\datasetheader{music-blues-reviews}}, column type=C{1.3cm}},
        columns/vegas-bars-reviews/.style={column name={\datasetheader{vegas-bars-reviews}}, column type=C{1.3cm}},
        columns/semantic-scholar-coauth-sample/.style={column name={\datasetheader{semantic-scholar-coauth-sample}}, column type=C{3.2cm}}
    }
}

\pgfplotstablenew[
    columns={AlgorithmKey,Algorithm},
    create on use/AlgorithmKey/.style={create col/set list={baseline_majority,cell2vec,celldiff2vec,deepcell,hoglee,hole,ComplexHeat,ComplexNetMF,ComplexRandNE,ComplexWalklets}},
    create on use/Algorithm/.style={create col/set list={\texttt{baseline\_majority},\texttt{cell2vec},\texttt{celldiff2vec},\texttt{deepcell},\texttt{hoglee},\texttt{hole},\texttt{ComplexHeat},\texttt{ComplexNetMF},\texttt{ComplexRandNE},\texttt{ComplexWalklets}}}
]{10}{\edgeclassificationwidetable}
\pgfplotsinvokeforeach{algebra-questions,cooking,geometry-questions,madison-restaurant-reviews,MAG-10,music-blues-reviews,vegas-bars-reviews}{
    \pgfplotstablecreatecol[create col/assign/.code={\setedgeclassificationresultcell{#1}}]{#1}{\edgeclassificationwidetable}
}

\pgfplotstablenew[
    columns={AlgorithmKey,Algorithm},
    create on use/AlgorithmKey/.style={create col/set list={baseline_mean,cell2vec,celldiff2vec,ComplexHeat,ComplexNetMF,ComplexRandNE,ComplexRep,ComplexWalklets,deepcell,hoglee,hole}},
    create on use/Algorithm/.style={create col/set list={\texttt{baseline\_mean},\texttt{cell2vec},\texttt{celldiff2vec},\texttt{ComplexHeat},\texttt{ComplexNetMF},\texttt{ComplexRandNE},\texttt{ComplexRep},\texttt{ComplexWalklets},\texttt{deepcell},\texttt{hoglee},\texttt{hole}}}
]{11}{\edgeregressionwidetable}
\pgfplotstablecreatecol[create col/assign/.code={\setedgeregressionresultcell{semantic-scholar-coauth-sample}}]{semantic-scholar-coauth-sample}{\edgeregressionwidetable}

\begin{table}[H]
    \centering
    \caption{%
        Edge classification accuracy values.
        Missing values indicate algorithm-dataset combinations that reached the one-hour timeout described in \Cref{section:experiments}.
    }%
    \label{table:edge-classification-results}
    \pgfplotstabletypeset[
        result table setup,
        columns={Algorithm,algebra-questions,cooking,geometry-questions,madison-restaurant-reviews,MAG-10,music-blues-reviews,vegas-bars-reviews},
    ]{\edgeclassificationwidetable}
\end{table}

\begin{table}[H]
    \centering
    \caption{Edge regression mean squared error values.}%
    \label{table:edge-regression-results}
    \pgfplotstabletypeset[
        result table setup,
        columns={Algorithm,semantic-scholar-coauth-sample},
    ]{\edgeregressionwidetable}
\end{table}

\end{document}